%% file: main.tex
\documentclass{article}

\PassOptionsToPackage{numbers, compress}{natbib}

 \usepackage[preprint]{main}

\usepackage[utf8]{inputenc} 
\usepackage[T1]{fontenc}    
\usepackage{hyperref}       
\usepackage{url}            
\usepackage{booktabs}       
\usepackage{amsfonts}       
\usepackage{nicefrac}       
\usepackage{microtype}      
\usepackage{xcolor}         

\usepackage{algorithm}
\usepackage{algorithmic}
\usepackage{amsmath}
\usepackage{booktabs}
\usepackage{multirow}
\usepackage{graphicx} 

\usepackage{newfloat}
\usepackage{listings}
\usepackage{placeins}
\usepackage{tcolorbox}
\title{Optimizing Prompt Sequences using Monte Carlo Tree Search for LLM-Based Optimization}

\author{
    Fei Xu Yu\\
    The George Washington University\\
    \texttt{fxyu@gwu.edu}\\
    \And
    Gina Adam\\
    The George Washington University\\
    \texttt{ginaadam@gwu.edu}
    \And
    Nathaniel D. Bastian\\
    United States Military Academy\\
    \texttt{nathaniel.bastian@westpoint.edu}\\
    \And
    Tian Lan\\
    The George Washington University\\
    \texttt{tlan@gwu.edu}
}

\begin{document}

\maketitle

\input{01abstract}

\input{02intro}

\input{03relatedwork}

\input{03.5prelim}

\input{04methodology}

\input{05experiment}

\input{06conclusion}

\medskip

{
\small
\bibliography{ref}
\bibliographystyle{unsrtnat} 
}

\input{appendix}

\end{document}

%% file: 01abstract.tex
\begin{abstract}
Mathematical optimization is a core tool in engineering, operations research, and data science,  but modeling real problems remains difficult and time-consuming. In practice, an expert must work with clients to translate informal domain descriptions into precise objectives, constraints, and solver-ready code; small mistakes in this process can lead to infeasible, nonconvex, or suboptimal formulations, and the iterative process is often slow. Large language models (LLMs) could automate this translation from natural language to optimization code, but standard prompting often fails on realistic tasks: it struggles with long, structured task descriptions, mishandles constraints, and produces infeasible, non-convex, or non-executable programs.
We introduce Monte Carlo Tree Search for Reasoning Decomposition Plans (MCTS-RDP), which performs pre-hoc test-time scaling by searching over decomposition plans and their resulting programs before committing to the final output. MCTS-RDP treats plan construction as a sequential decision problem: given a problem description, it decomposes the task descriptions into logical components and generates candidate plan steps for each component. Monte Carlo Tree Search (MCTS) then explores and refines Reasoning Decomposition Plans (RDPs) prior to code synthesis, balancing exploration and exploitation across alternative plan sequences. Each candidate plan is rolled out into executable code, executed, and scored with an execution-grounded evaluator measuring executability, constraint satisfaction, and solution quality; this feedback is backpropagated through the search tree to guide subsequent search.
Across wireless optimization tasks and the NL4OPT and NLP4LP benchmarks, MCTS-RDP consistently outperforms strong baselines, including one-shot prompting, Chain-of-Thought prompting, Self-Refine, and OPTIMUS, improving execution success, reward, stability, and optimality attainment.
\end{abstract}

%% file: 02intro.tex
\section{Introduction}

Mathematical optimization underpins many decision-making pipelines in engineering, operations research, and control. While modern solvers for linear, convex, and mixed-integer problems are highly mature, formulating the right model is still a major bottleneck. Typically, an operations research expert must work closely with a client to convert informal descriptions of costs, capacities, and business rules into precise decision variables, objectives, and constraints. This translation process is slow, requires specialized expertise, and is brittle: small modeling errors can lead to infeasible or non-convex problems, or to solutions that do not accurately reflect the original client intent. These limitations naturally raise the question of whether part of the modeling burden can be offloaded to automated tools. Large language models (LLMs) have demonstrated strong capabilities in generating code from natural language \cite{jiang2024surveylargelanguagemodels}, offering an appealing possibility: can a machine read a natural language description of an optimization problem and automatically produce mathematically correct, solver-ready code that faithfully encodes the underlying domain constraints and objectives?

However, current practices still fall short of this vision. LLMs are often used in one-shot or few-shot settings, where generation proceeds largely in a single pass. For optimization modeling tasks that require long-horizon reasoning and tight structural validity, this can yield programs that are syntactically plausible but logically inconsistent or solver-incompatible: constraints may be omitted or mistranslated, and non-convex expressions can violate modeling assumptions. These failure modes reflect the challenges of maintaining coherent, multi-step representations under autoregressive generation \cite{Li_2022,almorsi2025guidedcodegenerationllms}.

A natural response is to allocate additional inference-time computation. Recent pre-hoc test-time scaling methods spend compute before committing to a final output by explicitly exploring and evaluating intermediate reasoning. Some search over alternative reasoning branches in a solution space (e.g., Tree-of-Thought–style inference that expands and evaluates candidate branches) \cite{yao2023treethoughtsdeliberateproblem}, while others invest compute into generating higher-level guidance that steers downstream generation (e.g., RLAD) \cite{qu2025rladtrainingllmsdiscover}. We build on this pre-hoc perspective, but target a different planning object that is central to optimization modeling: the sequence of intermediate formulation decisions that determine feasibility and solver compatibility. Instead of only branching over complete candidate programs or reasoning traces, we reason over and prioritize decomposition plans, which are different ways to break the problem description into an ordered plan and realize objectives and constraints into solver-compatible code. In this setting, allocating inference-time computation to the formulation/planning space can be more effective than spending the same budget only at the final-solution stage.
For completeness, many widely used scaling strategies are post-hoc, generating multiple full candidates and selecting among them (e.g., Best-of-$N$); while effective in some cases, they largely treat early formulation choices as fixed.

In this work, we propose a \emph{pre-hoc} planning approach for optimization code synthesis that searches over how a natural-language optimization problem is structured for the model before selecting the final program. We introduce \textbf{Monte Carlo Tree Search for Reasoning Decomposition Plans (MCTS-RDP)}, a neural-symbolic framework that treats \emph{Reasoning Decomposition Plan} (RDP) construction as a sequential decision process solved with Monte Carlo Tree Search (MCTS). An RDP is an ordered sequence of plan steps that specifies how to translate a problem's context, objective, and constraints into solver-ready code. Throughout, prompts are only the textual queries used to interact with the role-specific LLMs, whereas the objects optimized by MCTS are the RDPs. Given a natural language optimization problem, we decompose it into logical components (e.g., context, objective, constraints) using an LLM with a fixed template. For each component, candidate plan steps are generated and scored by an LLM-based evaluator. MCTS explores and refines alternative RDPs via a UCT-style rule, rolling each candidate plan out into a complete Python program that is executed and scored using execution-grounded feedback; rewards are backpropagated to guide future plan choices. In effect, MCTS-RDP uses the optimization problem itself as feedback to learn which decomposition plans most reliably produce correct, feasible, and high-quality optimization programs. 

We evaluate MCTS-RDP on three sets of tasks: (i--ii) two natural-language optimization benchmarks, NL4OPT and NLP4LP, containing 100 and 242 questions; and (iii) 100 wireless network optimization problems involving power allocation under SINR and interference constraints. To further test generality, we augment NL4OPT with structurally enhanced instances that preserve convexity but increase difficulty by expanding the number of decision variables, adding auxiliary parameters, and incorporating additional constraints. Across both OpenAI GPT models and Google Gemini models, MCTS-RDP consistently outperforms one-shot prompting and strong baselines. On NL4OPT, MCTS-RDP achieves a 100\% execution success rate with a reward of 9.56/10, low variance (SD 1.41), 100\% constraint satisfaction, and 92\% optimality attainment. On NLP4LP, it maintains 100\% execution success with a 9.47 reward, 97.52\% constraint satisfaction, and 78.10\% optimality attainment.

The methodological novelty of MCTS-RDP lies not in applying MCTS to an LLM in general, but in defining the search space over ordered formulation instructions before final code synthesis. Our main contributions are:
\begin{enumerate}
    \item We introduce a plan-level search formulation for optimization
    code synthesis. A Reasoning Decomposition Plan is defined as an
    ordered sequence of formulation instructions associated with decision
    variables, objectives, constraints, and solver-compatible
    transformations.

    \item We develop an execution-grounded MCTS pipeline that integrates
    structured decomposition, plan-step generation, context-aware code
    synthesis, program execution, and reward backpropagation. This enables
    exploration and comparison of alternative formulation decisions before
    committing to the final program.

    \item We introduce a block-wise Search-then-Apply strategy that periodically performs plan search on a subset of instances and reuses the best block-local RDP on subsequent instances, thereby amortizing the cost of tree search.

    \item We evaluate MCTS-RDP across modified NL4OPT, NLP4LP, and wireless
    resource-allocation problems using multiple LLM backends. The
    evaluation considers executability, constraint satisfaction,
    optimality attainment, reward stability, token usage, and end-to-end
    inference time.
\end{enumerate}

%% file: 03relatedwork.tex
\section{Related Work}
\label{sec:related-work}
Recent advances in LLMs have enabled strong performance across a wide range of tasks, from natural language processing to code generation and high-level reasoning. However, applying LLMs to structured problem-solving, particularly tasks involving multi-step long-horizon planning and decision-making, remains a significant challenge. This has motivated research on improving LLM performance through better problem framing and decomposition, integration with symbolic planning techniques, and hybrid frameworks that combine learning-based and search-based methods. In this section, we review related work across four key areas.

\subsection{Large Language Models for Code Generation and Structured Problem Solving}
LLMs have evolved human-computer interaction by allowing more natural, context-aware communication. Studies have shown that LLMs can improve user productivity in domains such as software development, legal analysis, and scientific research by offering explanations and justifications that align with human reasoning \cite{openai2024gpt4technicalreport,geminiteam2024geminifamilyhighlycapable,zhang2025learning,ravari2024adversarial,zhang2024modeling}. A complementary line of work studies how factual knowledge is internally represented and shared across languages in LLMs, including evidence of language-agnostic "knowledge neurons" \cite{CAO2026104490}.
In the context of code generation, LLMs have been widely applied to tasks such as synthesis, translation, and bug repair \cite{chen2021evaluatinglargelanguagemodels, nijkamp2023codegenopenlargelanguage}. Recent empirical evidence also suggests that ChatGPT can perform useful code refactoring and documentation generation, although its reliability and behavior preservation are still not guaranteed in all cases \cite{DEPALMA2024123602}. While LLMs can generate correct code snippets in constrained settings, they often struggle with multi-step reasoning tasks or problems requiring precise constraint satisfaction, such as numerical optimization. Several studies \cite{zelikman2022starbootstrappingreasoningreasoning,madaan2023selfrefineiterativerefinementselffeedback,wei2023chainofthoughtpromptingelicitsreasoning,zhang2025network} have proposed methods to decompose problems into smaller subproblems or use natural language feedback to improve generation quality. However, these methods typically rely on hand-crafted chains of thought or heuristic reasoning templates, and they do not systematically optimize the sequence of reasoning steps for executability and solution quality.


\subsection{Integrations of LLMs with MCTS}
Recent works \cite{zhang2024accessinggpt4levelmathematical, zhang2024restmctsllmselftrainingprocess, zheng2025montecarlotreesearch,li2024rethinkmctsrefiningerroneousthoughts,lu2025lsrmctsalleviatinglongrange,zhang2025lipschitz} have explored using LLMs with MCTS \cite{Chaslot_Bakkes_Szita_Spronck_2021} to solve more complex, multi-step problems because MCTS is proven effective in sequential decision-making problems. However, existing approaches either focus on generating heuristic-based code or target relatively simple tasks where producing a correct answer suffices. These methods often overlook cases where LLMs generate correct but suboptimal solutions.

Existing LLM--MCTS approaches have demonstrated that tree search can
improve multi-step reasoning and code generation. The novelty of
MCTS-RDP therefore does not lie in applying MCTS to an LLM in general,
but in defining the search space at the level of formulation plans for
optimization code synthesis. In many existing LLM tree-search methods,
nodes represent general reasoning steps, partial answer trajectories,
partial programs, or complete solution candidates. In contrast, each
node in MCTS-RDP represents a formulation instruction associated with
a semantic component of the optimization problem, such as the decision
variables, objective, or constraints. A root-to-leaf path consequently
represents an ordered plan for translating the natural-language problem
into solver-compatible code.

This distinction allows the search to compare and revise early modeling
decisions before committing to the final program. This is particularly
important in optimization code synthesis because an incorrect variable
definition or constraint transformation may
propagate through the remaining generation process and may not be
recoverable through final-answer reranking. MCTS-RDP therefore uses
execution-grounded feedback to guide the search over the formulation
decisions that produce the final program.

\subsection{LLMs for Planning, Robotics, and Decision-Making}
LLMs have shown promising zero-shot generalization capabilities in structured reasoning and planning, particularly in generating high-level task sequences for complex decision-making processes\cite{guo2023advantage,wu2023planeliminatetrack,qiao2024br}. More recent work has also proposed multi-agent planning frameworks that explicitly target task-adaptive planning and execution under resource constraints \cite{ZHANG2026129423}. This paper \cite{yao2023reactsynergizingreasoningacting} studied the synergy between reasoning and acting of LLMs. Several papers studied ways to translate human instructions into executable task plans \cite{liu2023llmpempoweringlargelanguage,hazra2024saycanpayheuristicplanninglarge}. This paper \cite{vemprala2023chatgptroboticsdesignprinciples} studied the application of ChatGPT for robotics applications. Beyond robotics, LLMs have been applied to multi-agent coordination \cite{wang2024describeexplainplanselect,zou2024distributed,yang2025jamming,jing2024byzantine}, policy optimization \cite{shinn2023reflexionlanguageagentsverbal}, and knowledge retrieval \cite{xu2023rewoodecouplingreasoningobservations, zhao2024expelllmagentsexperiential,zhang2024distributed}. However, LLMs struggle with long-term coherence in planning tasks and with translating problem descriptions into precise objectives and constraints. Recent work further highlights that maintaining belief consistency is itself a core challenge for LLM agents in multi-turn decision-making settings \cite{HE2026129273}. This concern connects to broader efforts on making AI systems more predictable and reliable for decision-making settings \cite{ZHOU2026104491}. Related analyses also show that LLM reasoning can be sensitive to framing in complex decision contexts (e.g., moral and legal judgments) \cite{ALMEIDA2024104145}. Recent works explore integrating LLMs with constraint programming \cite{szeider2024mcpsolverintegratinglanguagemodels} and optimization frameworks \cite{huang2024largelanguagemodelmeets} to improve planning robustness. 

\subsection{Pre-hoc vs Post-hoc Test-time Scaling}

Recent advances in \emph{test-time scaling} allocate additional computation at inference time to improve reasoning and code generation quality. Pre-hoc approaches spend compute before producing the final output by exploring and evaluating intermediate reasoning artifacts. Some scale inference in the solution space by exploring multiple intermediate reasoning branches prior to committing to an answer, as in Tree-of-Thought \cite{yao2023treethoughtsdeliberateproblem} and Forest-of-Thought \cite{bi2025forestofthoughtscalingtesttimecompute}. Others invest compute into generating higher-level guidance that steers subsequent generation; for example, RLAD shows that allocating test-time compute toward producing such guidance can be more effective than simply sampling more solutions \cite{qu2025rladtrainingllmsdiscover}.

In contrast, \emph{post-hoc} scaling methods apply additional computation after candidates have been generated. The model produces multiple complete candidate programs under a largely fixed problem description, and a secondary mechanism then selects, aggregates, or reranks candidates. Examples include Best-of-$N$ sampling, which chooses the highest-scoring candidate among multiple generations, and OPTIMUS \cite{ahmaditeshnizi2025optimus03usinglargelanguage}, which improves outcomes via generation-and-selection using an external evaluator or selection rule. While these approaches can yield clear gains, they primarily improve performance by selecting among final candidates and may be less effective when errors arise from early modeling choices that propagate into infeasibility or solver-incompatible formulations.

%% file: 03.5prelim.tex
\section{Preliminaries}
In this section, we introduce the basic concepts used in our framework: mathematical optimization problems, LLMs for code generation, MCTS, and the Upper Confidence bound applied to Trees (UCT) rule that we use for plan step selection.

\subsection{Mathematical Optimization Problems}
We consider standard optimization problems of the form
\begin{equation}
\begin{aligned}
    \min_{x \in \mathcal{X}} \quad & f(x) \\
    \text{subject to} \quad 
    & g_i(x) \le 0, \quad i = 1,\dots,m, \\
    & h_j(x) = 0, \quad j = 1,\dots,p.
\end{aligned}
\end{equation}

where $x \in \mathbb{R}^n$ denotes the decision variables, $f$ is the objective function, and $g_i$, $h_j$ encode inequality and equality constraints, respectively. In many of our benchmarks (e.g., NL4OPT, NLP4LP, and the wireless tasks), $f$ and $g_i$ are linear or convex, so the problems can be solved reliably by off-the-shelf solvers once a correct formulation is available.

In practice, however, the difficulty lies not in solving the optimization problem, but in modeling it correctly. Given a natural language description, the modeler must:
(i) identify the decision variables (e.g., powers, flows, or allocation levels),
(ii) construct the objective (e.g., minimize power, cost, or time), and
(iii) encode all constraints (e.g., capacity, SINR, budget, logical relations) in a solver-compatible form.
Our goal is to automate this mapping from natural language descriptions to executable optimization code, while preserving feasibility and optimality whenever possible.

\subsection{Large Language Models for Code Synthesis}
We treat an LLM as a conditional generative model that maps a natural language prompt $P$ to a distribution over token sequences, $Y \sim p_\theta(\cdot \mid P),$
where $\theta$ denotes the fixed model parameters. In our setup, we use RDPs to structure the interaction with the LLM.
Concretely, each plan step is instantiated as a natural-language prompt that describes a specific
component of the optimization task (e.g., context, objective, or constraint block). In later stages, these prompts may also include partial code or structural hints accumulated so far. The LLM is then asked to generate solver-ready Python programs using an appropriate optimization library (e.g., CVXPY, PuLP,
Pyomo, OR-Tools).

We assume the LLM is a black box: we can query it multiple times with different reasoning decompositions or reasoning decomposition steps, but we do not modify or fine-tune the model itself. The quality of the generated code depends sensitively on how the problem is described. Poor decompositions may result in:
(i) missing or incorrect constraints,
(ii) non-convex formulations that violate solver assumptions, or
(iii) code that fails to execute.
This motivates our focus on reasoning decomposition planning as a first-class object of optimization.

\subsection{Monte Carlo Tree Search }
MCTS is an iterative tree search algorithm that alternates between exploration and exploitation by simulating trajectories in a search tree. A generic MCTS iteration consists of four phases:
\begin{enumerate}
    \item \textbf{Selection}: starting from the root node, iteratively select child nodes according to a tree policy (e.g., UCT) until a leaf node is reached.
    \item \textbf{Expansion}: if the leaf node is non-terminal and not fully expanded, add one or more new child nodes corresponding to previously unexplored actions.
    \item \textbf{Simulation (rollout)}: from the selected or newly expanded node, simulate a trajectory to a terminal state using a default policy, and obtain a scalar reward $R$ (typically assumed to be bounded).
    \item \textbf{Backpropagation}: propagate the reward $R$ back up the tree, updating visit counts and value estimates for all nodes along the path from the leaf to the root.
\end{enumerate}

To balance exploration and exploitation during selection, we adopt the UCT rule. For a node $v$ with parent $p(v)$, each child $u \in \mathcal{C}(v)$ is assigned a score:
\begin{equation}
    \mathrm{UCT}(u)
    =
    \hat{r}(u)
    +
    c \sqrt{\frac{\log N(v)}{N(u)}},
    \label{eq:uct}
\end{equation}
where $\hat{r}(u)$ is the empirical mean reward at $u$, $N(v)$ and $N(u)$ are the visit counts of the parent and child, respectively, and $c > 0$ is an exploration constant. The first term favors children with high observed rewards (exploitation), while the second term favors less-visited children (exploration).

\begin{figure*}[t]
\centering
\includegraphics[width=1\textwidth]{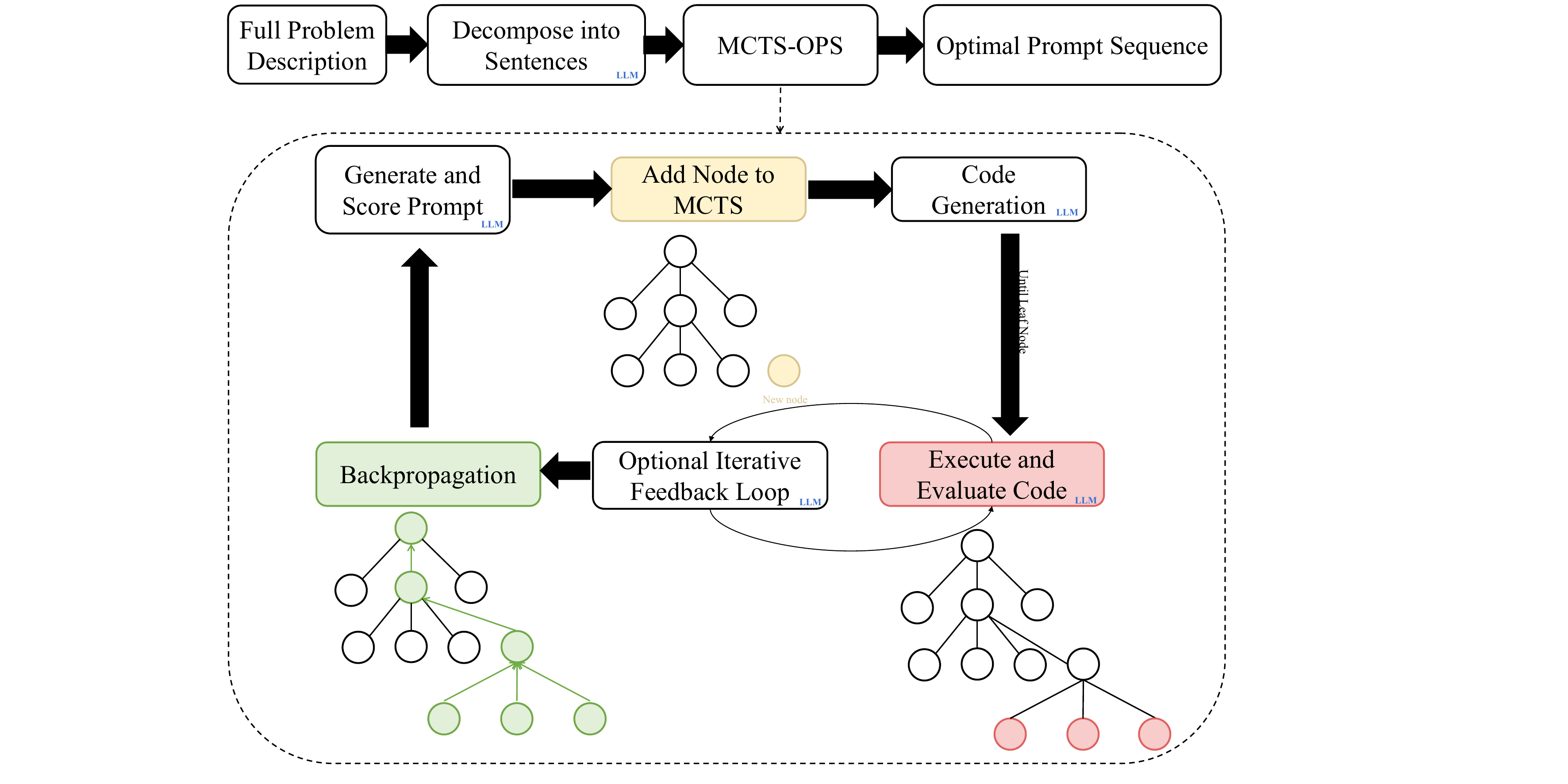}
\caption{Overview of \textbf{MCTS-RDP}. Given a textual optimization problem, the input is decomposed into structured logical components using an LLM
with a fixed template (context, objective, constraints). For each component, the method proposes candidate plan steps (i.e., elements of a RDP), scores them with an LLM-based evaluator, and organizes them in an MCTS tree. Traversing a root-to-leaf path yields a complete RDP, which is rolled out into a full Python program. The program is executed and evaluated for executability, feasibility, and optimality; the resulting reward is backpropagated to update search statistics. An optional iterative refinement loop performs bounded local repairs to improve faulty scripts.}
\label{fig:1}
\end{figure*}

%% file: 04methodology.tex
\section{Methodology}

\begin{algorithm}[!h]
\caption{MCTS-RDP}
\label{MCTSRDP}
\begin{algorithmic}[1]
\REQUIRE Problem description $\mathcal{X}$, total simulations $N$
\STATE $T \gets$ Initialize MCTS tree with root node
\STATE $S \gets$ Decompose($\mathcal{X}$) \COMMENT{structured components / atomic elements}

\FOR{$i = 1$ to $N$}
    \STATE $node \gets$ root of $T$
    \STATE $code \gets$ empty script
    \STATE $path \gets$ empty list

    \FOR{each component $s \in S$}
        \IF{$node$ has children}
            \STATE $node \gets$ SelectChildUCT($node$)
            \STATE $d \gets node.\text{plan\_step}$
        \ELSE
            \STATE $d \gets$ ProposePlanStep($s$) \COMMENT{LLM proposes a plan step}
            \STATE $score \gets$ ScorePlanStep($d \mid s, \mathcal{X}$) \COMMENT{evaluator score in $[0,10]$}

            \IF{SiblingWithScoreExists($node$, $score$)}
                \STATE $node \gets$ ReuseSibling($node$, $score$)
                \STATE $d \gets node.\text{plan\_step}$
            \ELSE
                \STATE $node \gets$ ExpandNode($node$, $d$, $score$)
            \ENDIF
        \ENDIF

        \STATE $code \gets$ UpdateProgram($d$, $code$) \COMMENT{context-aware regeneration}
        \STATE Append($path$, ($d$, $score$))
    \ENDFOR

    \STATE $reward \gets$ EvaluateCode($code$)
    \IF{$reward < \tau$}
        \STATE $(code, reward) \gets$ OptionalRefine($code$, $reward$)
    \ENDIF
    \STATE Backpropagate($T$, $node$, $reward$)
\ENDFOR
\end{algorithmic}
\end{algorithm}

This section presents the complete MCTS-RDP framework. We first formalize Reasoning Decomposition Plans and describe LLM-guided decomposition and context-aware code generation. We then introduce the execution-grounded reward, explain its integration with MCTS, and describe the optional refinement and search--then--apply procedures. We proffer \textbf{Monte Carlo Tree Search for Reasoning Decomposition
Plans (MCTS-RDP)}, a neural-symbolic framework that formulates Reasoning
Decomposition Plan construction as a sequential decision process. For an
ordered sequence of problem components $S=(s_1,\ldots,s_T)$, an RDP is
$\pi=(d_1,\ldots,d_T)$, where each $d_t$ is a formulation instruction
associated with component $s_t$. A partial RDP represents the plan-level search state, and selecting the
next instruction constitutes an MCTS action. Each root-to-leaf path therefore represents a complete formulation plan rather
than a completed solution or an unrestricted reasoning trace. The resulting
program is executed, and its execution-grounded reward is backpropagated
through the plan nodes.
The framework integrates LLM-guided decomposition, context-aware code synthesis, structured execution-grounded reward evaluation, and UCT-guided search. We provide the full system prompt templates for each role (decomposer, planner, code synthesizer, evaluator) in~\ref{app:system-prompts}, and the complete procedure is given in Algorithms~\ref{MCTSRDP} and~\ref{alg2}.

\subsection{LLM-Guided Decomposition and Plan-Step Generation}
\label{sec:decomposition-planstep}

Let a natural language optimization problem be denoted by $\mathcal{X} = (\mathcal{C}, \mathcal{O}, \mathcal{K})$, where $\mathcal{C}$ is contextual description, $\mathcal{O}$ is the objective, and $\mathcal{K}$ is the set of constraints. We use an LLM to produce a structured decomposition: $
S = \text{Decompose}(\mathcal{X}) = ( s_1, s_2, \dots, s_T )$,
where each $s_t$ corresponds to a semantically atomic element of the problem description. For each $s_t$, an LLM proposes a candidate step instruction: 
$
d_t = \text{ProposePlanStep}(s_t)
$. A plan step is a natural-language instruction that specifies what to encode in the program at this stage. An LLM generates a score reflecting semantic alignment, clarity, and usefulness for correct code synthesis: $
Score = \text{ScorePlanStep}(d_t \mid s_t, \mathcal{X}) \in [0,10]$.
Each tree node stores a $(d_t,\mathrm{Score})$ pair. To control the branching factor, we use a lightweight score-based sibling-reuse heuristic. If a newly proposed plan step receives the same discrete evaluator score as an existing sibling, we reuse the existing sibling rather than creating an additional branch. This rule is intended only as a computational heuristic for limiting tree growth. Equal evaluator scores do not imply semantic equivalence: distinct plan steps may receive the same score, while semantically similar steps may receive different scores. We therefore refer to this mechanism as \emph{score-based sibling reuse} rather than semantic deduplication.

\subsection{Context-Aware Code Generation}
Given a partial RDP $\pi = (d_1, \dots, d_t)$, MCTS-RDP conditions an LLM on both the current plan step $d_t$ and the current cumulative code  $P_{t-1}$ to produce an updated program: $P_t = \text{UpdateProgram}(d_t \mid P_{t-1}).$
We implement this as incremental regeneration: the LLM uses $P_{t-1}$ as context but outputs a complete updated script $P_t$, which helps maintain consistent variable declarations, constraints, and solver calls across steps. 

The current implementation does not apply a separate hard validity classifier or action mask before adding a candidate plan step to the search tree. Instead, invalidity is handled through context-aware code generation and execution-grounded feedback. Because each update is conditioned on the cumulative program $P_{t-1}$, the code-generation model can preserve previously introduced variable names, solver choices, constraints, and program structure. 
If a plan step produces syntactically incompatible code or a runtime failure, the completed rollout receives a reward of $-1$. If the program executes but is logically inconsistent with the problem or previously encoded constraints, the evaluator assigns a lower reward based on its assessment of constraint satisfaction and objective quality. This reward is backpropagated through the corresponding branch, making that branch less likely to be selected in later simulations. When the reward falls below the refinement threshold, the optional bounded refinement procedure may also attempt a local repair.

\subsection{Reward Evaluation}
After a complete program $P$ is produced, it is executed in a Python environment. A secondary LLM evaluator receives the original problem description and the complete execution trace, including standard output and error output. When a reference numerical solution is available, it is also provided to the evaluator together with the numerical matching tolerance. The evaluator operates only after program execution. 

Using a fixed prompt, the evaluator returns a structured JSON object containing a scalar reward $R(P)\in[-1,10]$, a textual explanation, an execution-success indicator, an evaluator-assessed constraint-satisfaction indicator, and an indicator of whether the numerical output matches the reference solution. Programs that fail because of a syntax error, runtime error, or timeout receive a reward of $-1$. Successfully executed programs receive a nonnegative score based on output completeness, constraint satisfaction, objective quality, and, when available, numerical agreement with the reference solution.
 
The scalar reward is used directly in MCTS selection and backpropagation, whereas execution success, constraint satisfaction, and optimality attainment are reported separately as evaluation metrics. Importantly, the reward is not computed from a manually specified weighted sum, and the prompt does not define fixed numerical weights or a fixed base-score decomposition for the individual criteria. Rather, the secondary LLM applies the same qualitative scoring rubric to every generated program. Consequently, the scalar reward is potentially stochastic even when the program and execution trace are held fixed. We therefore evaluate the repeatability and ranking stability of this evaluator separately in Section~\ref{sec:evaluator-reliability}. The complete evaluation prompt is provided in~\ref{app:system-prompts}.

\subsection{Monte Carlo Tree Search Integration}
MCTS-RDP follows the standard phases of selection, expansion, simulation, and backpropagation. Each node corresponds to a partial RDP, and a complete path corresponds to a full RDP.

During selection, the algorithm traverses from the root and chooses child nodes that maximize the UCT score: $
\text{UCT}(v) = \frac{Q(v)}{N(v)} + c \cdot \sqrt{\frac{\log N(p)}{N(v)}},
$
where $Q(v)$ is cumulative reward, $N(v)$ is visit count, $N(p)$ is the parent’s visit count, and $c > 0$ controls exploration.In expansion, a candidate plan step is proposed for the next unresolved problem segment and scored by the evaluator. If no existing sibling has the same discrete evaluator score, a new child node is created; otherwise, the existing same-score sibling is reused according to the score-based branching-control heuristic described above. In simulation, the selected RDP is rolled out into a full program, executed, and assigned a scalar reward. Finally, backpropagation propagates this reward from leaf to root, updating statistics so future search increasingly favors high-performing decomposition plans. Although the optional refinement loop may locally repair code, we still attribute the resulting reward to the RDP that produced the initial program, treating refinement as a bounded local repair rather than a new plan branch.

To better understand the theoretical foundation of our method, we examine its performance through the lens of regret minimization. At its core, MCTS-RDP traverses a decision tree where each node corresponds to a partial plan step in a generated sequence. Using the UCT rule, the algorithm selects actions that optimally balance exploration and exploitation. This decision process is inherently stochastic due to the nature of LLM outputs and reward evaluations. To ground our design in established theory and validate our confidence-based selection, we adapt classical UCT analysis to our continuous, bounded reward setting. In our analysis, we assume rewards are bounded in a fixed interval, $R(v) \in [-1,10]$, so that concentration inequalities (e.g., Hoeffding bounds) apply. This boundedness allows us to relate the number of times a suboptimal node is selected to the size of its suboptimality gap and to derive logarithmic-in-samples regret bounds analogous to classical UCT analyses, adapted to our LLM-guided setting. In the section below, we derive a regret bound tailored to MCTS-RDP, showing that the expected suboptimality diminishes logarithmically with the number of simulations, which is a desirable property for sample-efficient planning under uncertainty. 
\subsection{Theoretical Analysis: Regret Bounds for UCT in LLM-Guided Tree Search}

We analyze the behavior of our UCT-based Monte Carlo Tree Search framework in the context of LLM-guided code synthesis. Let each node $v$ in the tree represent a partial plan step. The search progresses by expanding nodes that optimize a UCT objective, balancing reward exploration and exploitation.

Let $X_{v,i} \in [-1, 10]$ denote the reward returned by the evaluator LLM after the $i$-th simulation from node $v$. Define the empirical mean as:
\[
\hat{\mu}_v = \frac{1}{N(v)} \sum_{i=1}^{N(v)} X_{v,i}
\]
where $N(v)$ is the number of visits to node $v$, and $\mu_v = \mathbb{E}[X_{v,i}]$ is the true expected reward. Assuming rewards are independent and bounded in range $R = 11$, we apply Hoeffding’s inequality:
\[
\Pr\left(|\hat{\mu}_v - \mu_v| \geq \epsilon\right) \leq 2 \exp\left(-\frac{2N(v)\epsilon^2}{R^2}\right)
\]

To analyze regret, suppose $a^*$ is the optimal child of a parent node $v$, and $u$ is a suboptimal child with a suboptimality gap $\Delta = \mu_{a^*} - \mu_u > 0$. The UCT score for node $u$ is:
\[
\text{UCT}(v \rightarrow u) = \hat{\mu}_u + c \sqrt{\frac{\log N(v)}{N(u)}}
\]

Node $u$ is chosen over $a^*$ only if:
\[
\hat{\mu}_u + c \sqrt{\frac{\log N(v)}{N(u)}} \geq \hat{\mu}_{a^*} + c \sqrt{\frac{\log N(v)}{N(a^*)}}
\]

Using a union bound and Hoeffding’s inequality on both arms, we get:
\[
\Pr(\text{select } u \text{ at time } t) \leq \Pr\left(\hat{\mu}_u \geq \mu_u + \frac{\Delta}{2}\right) + \Pr\left(\hat{\mu}_{a^*} \leq \mu_{a^*} - \frac{\Delta}{2}\right)
\]
\[
\leq 2 \exp\left(-\frac{2N(u)\Delta^2}{4R^2}\right) = 2 \exp\left(-\frac{N(u)\Delta^2}{2R^2}\right)
\]

Summing this probability over $T$ total steps yields a bound on the expected number of visits to suboptimal $u$:
\[
\mathbb{E}[N_T(u)] \leq \frac{2R^2 \log T}{\Delta^2} + C
\]

\textbf{Final Regret Bound.} The total regret over $T$ simulations is:
\[
\text{Regret}_T = \sum_{u \neq a^*} \Delta_u \cdot \mathbb{E}[N_T(u)]
\leq \sum_{u \neq a^*} \Delta_u \left( \frac{2R^2 \log T}{\Delta_u^2} + C \right) = \mathcal{O}(\log T)
\]

This shows that MCTS-RDP maintains the classical logarithmic regret bound in the number of simulations, even when rewards are real-valued, continuous, and bounded. This adaptation strengthens the theoretical foundation of using UCT with noisy LLM-based reward feedback.

\begin{algorithm}[h]
\caption{Optional Iterative Feedback Loop}
\label{alg2}
\begin{algorithmic}[1]
\REQUIRE Max retries $R$, reward threshold $\tau$

    \STATE $r \gets 0$
    \WHILE{$reward < \tau$ and $r < R$}
        \STATE $feedback \gets$ GetFeedback($code$, $reward$)
        \STATE $code \gets$ ReviseCode($code$, $feedback$)
        \STATE $reward \gets$ EvaluateCode($code$)
        \STATE $r \gets r + 1$
    \ENDWHILE

\end{algorithmic}
\end{algorithm}

\subsection{Iterative Refinement}
If the reward $R(P)$ assigned to a generated program falls below a threshold $\tau$, the evaluator produces feedback $f$ and the LLM revises:
\[
P' = \text{ReviseCode}(\pi \mid P, f), \quad \text{with } R(P') > R(P) \text{ desired},
\]
with a bounded retry budget $M$. This mechanism corrects local syntax errors without restarting the full search.

\subsection{Search--then--Apply Variant}
\label{sta}

Running full MCTS-RDP on every instance is computationally expensive because each search simulation requires multiple LLM calls for plan-step generation, code synthesis, execution, and evaluation. We therefore use a search-then-apply variant that periodically performs tree search and reuses the resulting formulation plan on subsequent instances.

For a benchmark containing $N$ instances, we divide the instances sequentially into approximately ten blocks, so that each block contains roughly $N/10$ instances. Within each full block, approximately $10\%$ of the instances are assigned to SEARCH mode. Because this quantity need not be integral, we take the ceiling. Thus, for a block of size $B$, the number of SEARCH instances is $ k_b = \left\lceil 0.1 B \right\rceil $. The first $k_b$ instances of each block run full MCTS-RDP. After each SEARCH instance, the highest-reward completed rollout observed so far within that block is retained as the reusable exemplar, with lower token usage used to break reward ties. The retained exemplar is a compact three-step representation of the selected RDP trajectory rather than the complete search tree.

The remaining instances in the block are processed in APPLY mode. Tree search is disabled, and code is generated in a single pass conditioned on the retained block exemplar. 

For modified NL4OPT ($N=100$), the benchmark is divided into blocks of 10 instances, with one SEARCH instance per block. For NLP4LP ($N=242$), the corresponding full-block size is 24, giving $k_b=\lceil 0.1\times24\rceil=3$ SEARCH instances per full block. The 242-instance benchmark contains ten complete blocks plus a final partial block of two instances. Because both remaining instances occupy SEARCH positions in the new block, the archived GPT-5 run contains 32 SEARCH and 210 APPLY instances overall, corresponding to a realized search fraction of $32/242=13.2\%$.

%% file: 05experiment.tex
\section{Experiment}

We evaluate the performance of our proposed method, \textbf{MCTS-RDP}, on a variety of natural-language optimization problems spanning three datasets: 100 modified NL4OPT tasks, 242 NLP4LP tasks, and 100 wireless resource allocation tasks. We augment the standard NL4OPT benchmark by expanding the decision variable set, introducing layered and interdependent constraints, and enriching contextual descriptions. These modified NL4OPT problems preserve convexity but substantially increase structural difficulty, creating a more challenging test case for LLM-based optimization code synthesis. NLP4LP provides an additional LP-focused benchmark, enabling a broader evaluation of robustness across linear programming instances. To test generality beyond LP-style problems, we also construct wireless resource allocation tasks that model realistic scenarios such as power minimization under Signal-to-Interference-plus-Noise Ratio (SINR) constraints and user interference, where correct formulation requires careful constraint modeling to ensure feasibility and convexity.

Our goal is to assess how effectively MCTS-RDP can guide LLMs to generate correct, efficient, and generalizable Python programs that solve these problems. To comprehensively benchmark our method, we compare MCTS-RDP against several LLM-based baselines. Specifically, we consider one-shot prompting using GPT-3.5-Turbo, GPT-4 \cite{openai2024gpt4technicalreport}, GPT-5 \cite{singh2025openaigpt5card}, and Gemini-2.5-Pro \cite{comanici2025gemini25pushingfrontier}; Chain-of-Thought (CoT) \cite{wei2023chainofthoughtpromptingelicitsreasoning}, which encourages step-by-step reasoning before generating solver code; Self-Refine \cite{madaan2023selfrefineiterativerefinementselffeedback}, where an initial solution is iteratively revised using self-generated feedback; and OPTIMUS \cite{ahmaditeshnizi2025optimus03usinglargelanguage} as a representative generation-and-selection pipeline. Across all methods, we measure execution success rate, average reward and standard deviation, constraint satisfaction rate, optimality attainment, average token usage, and average end-to-end inference time under the same evaluation protocol. A generated solution is counted as attaining optimality when the evaluator's judgment indicates that the extracted numerical result matches the reference solution within the specified tolerance. Feasible but suboptimal solutions are therefore counted as constraint-satisfying but not optimal. We separately assess agreement between this evaluator judgment and a deterministic numerical comparison in Section~\ref{sec:evaluator-reliability}.

For MCTS-RDP, we fix the maximum tree depth and the UCT exploration constant. We also report token usage statistics for each method and analyze the trade-off between performance and cost in section~\ref{sec:cost}. Together, these metrics capture improvements in executability, feasibility, and optimality enabled by plan search, beyond what single-shot generation achieves. We also conduct ablation studies to isolate the contributions of tree-based plan search and plan decomposition. We report the average end-to-end inference time, which is measured from the submission of the natural-language problem description until the final evaluated program and reward are returned. The measurement includes all LLM calls, plan generation, code synthesis, tree search, evaluator calls, optional refinement, and Python program execution. All methods were run sequentially under the same experimental environment. For the search-then-apply variant, the reported average end-to-end time includes both SEARCH and APPLY instances, with total time divided by the number of benchmark instances. All experiments are conducted on a Linux machine with AMD EPYC 7513 32-Core Processor CPU and an NVIDIA RTX A6000 GPU, implemented in Python 3.

The remainder of this section first describes the benchmarks and evaluation protocol, then reports the main results and baseline comparisons. We subsequently analyze computational cost, present the ablation study, and examine the remaining failures on the hard wireless benchmark.

\subsection{Problem Setup}

We evaluate all methods across 442 natural language optimization tasks: 100 modified NL4OPT problems, 242 NLP4LP problems, and 100 wireless communication problems. Each problem is described entirely in natural language and requires the generation of correct, executable Python code that produces valid numerical solutions. The code must solve the problem’s objective while satisfying all constraints. We ensure a consistent evaluation setup by providing all models with the exact same problem descriptions. We define the execution success rate as the generation of syntactically correct and executable code that returns a valid solution; we additionally measure constraint satisfaction, optimality attainment, token usage, and end-to-end inference time.

\textbf{Modified NL4Opt Benchmarks}
To further test generality and robustness, we augment the standard NL4Opt benchmark with 100 structurally enhanced tasks. The original NL4Opt problems are typically concise, with minimal variable sets and clearly defined objectives and constraints. In contrast, our modified NL4Opt instances are designed to better simulate real-world technical problem descriptions by introducing several key enhancements. First, we increase the dimensionality of the optimization problems by adding more decision variables (e.g., multiple products, assets, or agents) and introducing auxiliary parameters (such as penalties, weights, or priorities). These additions create richer problem structures and more intricate decision spaces, testing whether LLMs can scale reasoning decomposition to larger mathematical programs. Second, we incorporate a greater number and diversity of constraints, including coupling constraints that link multiple variables together (e.g., “if X is chosen, then Y must be above a threshold”) and global constraints involving sums, products, or nested conditions. These new constraints simulate realistic operational rules and regulatory requirements often seen in applied optimization settings.

\textbf{NLP4LP Benchmarks}
NLP4LP contains 242 linear programming tasks described in natural language. This benchmark complements NL4OPT by emphasizing LP structure and testing whether methods can reliably translate descriptions into correct LP formulations and solver-ready code across a broader set of instances.

\textbf{Wireless Communication Benchmarks}
\label{sec:wireless-setup}
To better understand the performance of each method, we partition the 100 problems into two equally sized subsets: 50 easy problems and 50 hard problems, based on the structure and complexity of their constraint formulations. The easy problems are characterized by a single, straightforward constraint: bounding each user's transmit power within a specified range. These problems are relatively simple, as they require no intricate reasoning or compound constraint expressions. Notably, we found that even simple one-shot prompting methods with a single natural language prompt can frequently solve these tasks correctly, generating valid and feasible code in a majority of runs. In contrast, the hard problems introduce an additional non-trivial constraint: each user must satisfy a minimum SINR requirement. This makes the problem more realistic in wireless systems but also considerably more challenging for LLMs. The SINR constraint expression is a function where both the numerator and the denominator contain optimization variables, which are the transmit power terms for both the target user and the interfering user. This violates Disciplined Convex Programming (DCP) rules, which restrict operations to maintain convexity and feasibility guarantees in modeling tools like CVXPY. Formulating this correctly in code requires expressing the logic accurately while adhering to the domain's convexity rules. By deliberately separating the easy and hard problems, we can better demonstrate the advantages of our proposed MCTS-RDP approach by incrementally building and refining code with tree-based guidance and reward-driven feedback.

\noindent\fbox{%
    \parbox{0.97\linewidth}{%
        \textbf{Example Easy Network Optimization Problem:} Two users transmit to a base station. Each user has a channel gain to the base station (1.5 and 1.0, respectively) and experiences background noise of 1 Watt. The objective is to minimize the total transmit power of both users, subject to the constraint that each user's transmit power must be between 0 and 1 Watt.
    }%
}

\vspace{0.1in}

\noindent\fbox{%
    \parbox{0.97\linewidth}{%
        \textbf{Example Hard Network Optimization Problem:} A wireless communication network includes 2 mobile users and 1 base station. The objective is to minimize the total transmit power used by both users, while ensuring quality of service. Each user transmits to the base station and must achieve a minimum Signal-to-Interference-plus-Noise Ratio (SINR) of -1.5 dB. The SINR for each user depends on their respective channel gain to the base station, interference from the other users, and a background noise power of 1 Watt. The channel gain from user 1 to the base station is 3.0, and from user 2 to the base station is 2.0. Each user’s transmit power must lie between 0 and 2 Watt.
    }%
}

\textbf{Evaluation Protocol.}

For NL4OPT and NLP4LP, we use the search--then--apply procedure described in Section~\ref{sta}. The benchmark instances are processed in their dataset order and divided into approximately ten sequential blocks. Within each block, the first $k_b=\lceil 0.1B\rceil$ instances are assigned to SEARCH mode. The remaining instances are processed in APPLY mode using the highest-reward block exemplar identified during SEARCH. For modified NL4OPT, $B=10$ and $k_b=1$. For NLP4LP, $B=24$ and $k_b=3$; the final partial block yields 32 SEARCH and 210 APPLY instances overall. Complete experimental configuration, metric aggregation definitions, and
details of the released artifacts are provided in
\ref{app:reproducibility}.
For the wireless benchmarks, we report the full MCTS-RDP that performs tree search separately for every instance. Reference solutions are not included in the initial code-generation prompt and are used during post-execution evaluation to assess optimality. NL4OPT and NLP4LP provide reference solutions directly. For the wireless benchmark, we obtain reference solutions using a reference convex formulation and a standard optimization solver.

\FloatBarrier
\input{table2}
\input{table3}
\FloatBarrier

\textbf{Results on NL4OPT and NLP4LP}
Tables~\ref{tab:nl4opt} and~\ref{tab:nlp4lp} summarize performance on modified NL4OPT and NLP4LP using GPT-5 and Gemini-2.5-Pro. On modified NL4OPT, MCTS-RDP attains \textbf{100\%} execution success for both backends, with reward \textbf{9.56/9.59} (SD \textbf{1.41/1.46}), \textbf{100\%/98\%} constraint satisfaction, and \textbf{92\%/91\%} optimality, using \textbf{7,517/7,247} average tokens and \textbf{100.26/39.72}s average end-to-end inference time. On NLP4LP, MCTS-RDP again reaches \textbf{100\%} execution success with reward \textbf{9.47/9.51} (SD \textbf{1.37/1.59}), \textbf{97.52\%/97.52\%} constraint satisfaction, and \textbf{78.10\%/76.45\%} optimality, with \textbf{7,158/6,940} average tokens and \textbf{70.08/38.28}s average end-to-end inference time. Across both benchmarks, MCTS-RDP yields the strongest combination of executability, feasibility, and optimality among the reported methods.

\FloatBarrier
\input{table1}
\FloatBarrier
\textbf{Results on Wireless Benchmarks}
Table~\ref{table1} summarizes performance on 100 network optimization tasks (50 easy, 50 hard).
On the easy subset, MCTS-RDP achieves \textbf{98\%} execution success with reward \textbf{$9.11 \pm 0.43$} and \textbf{92\%} optimality.
Relative to one-shot baselines, this corresponds to a \textbf{+26} point gain in success rate over GPT-4, nearly \textbf{2}$\times$ higher reward, and \textbf{3.0}$\times$ lower reward variability.
Compared to stronger iterative baselines, MCTS-RDP delivers the highest reward stability and the best optimality rate.

On the hard subset, MCTS-RDP is the only method that reliably solves a substantial fraction of instances, reaching \textbf{70\%} success with reward \textbf{$6.75 \pm 3.89$}.
Moreover, MCTS-RDP raises optimality attainment from \textbf{near-zero} for all baselines to \textbf{66\%}.
These gains come with higher token usage, reflecting the additional inference-time planning required to preserve feasibility and solver compatibility in the hard setting.

\subsection{Comparison with Baselines}
MCTS-RDP is evaluated against one-shot generation, Chain-of-Thought, Self-Refine, and OPTIMUS. These baselines differ primarily in when they allocate test-time compute. One-shot methods commit to a single end-to-end generation; CoT adds an explicit reasoning trace but does not systematically revise the modeling plan; Self-Refine performs local iterative repairs after an initial draft; and OPTIMUS improves outcomes through repeated generation-and-selection with an external evaluator.
In contrast, MCTS-RDP allocates compute to plan search: it searches over alternative RDPs (sequences of plan steps for encoding context, objective, and constraints) and uses execution-grounded rewards to guide exploration.

Empirically, this leads to two consistent advantages. First, MCTS-RDP maintains perfect (or near-perfect) executability while improving feasibility: constraint satisfaction remains near \textbf{98--99\%} on both NL4OPT and NLP4LP. Second, it substantially increases solution quality beyond mere executability, attaining the highest optimality rates on both datasets (Tables~\ref{tab:nl4opt},~\ref{tab:nlp4lp}). On wireless tasks, the gains are most pronounced on the hard subset, where naive formulations often violate convexity requirements; plan search improves both success and reward by steering the model toward feasible SINR-compliant formulations. Our experiments reveal that baseline methods (one-shot prompting, and even CoT and Self-Refine) struggle consistently with these harder problems. The issues occur from two primary sources: (1) Syntax error: The LLMs often generate incomplete or incorrect expressions, such as using undefined variables or incorrect mathematical operations. (2) Infeasible solutions: Even when the generated code is syntactically correct, it often produces infeasible solutions due to incorrect constraint formulations or improper parameter settings. These failures highlight the limitations of naive prompting methods in translating complex natural language requirements into executable and correct optimization formulations.

\subsection{Evaluator Reliability and Ranking Stability} \label{sec:evaluator-reliability}

\begin{table}[t]
\centering
\small
\setlength{\tabcolsep}{4pt}
\begin{tabular}{lccc}
\toprule
Metric & Overall ($n{=}30$) & Optimal ($n{=}15$) & Suboptimal ($n{=}15$) \\
\midrule
\multicolumn{4}{l}{\textit{Within-evaluator Boolean agreement}} \\
Executed & $100.0\%$ & $100.0\%$ & $100.0\%$ \\
Constraint satisfied & $99.3\%$ & --- & --- \\
Ground-truth match & $100.0\%$ & $100.0\%$ & $100.0\%$ \\
\midrule
\multicolumn{4}{l}{\textit{Scalar reward variation}} \\
Mean within-trace SD & $0.27$ & $0.00$ & $0.54$ \\
Mean within-trace range & $0.69$ & $0.00$ & $1.38$ \\
Max within-trace range & $6.2$ & $0.0$ & $6.2$ \\
Traces crossing score $10$ & $2$ & $0$ & $2$ \\
\midrule
\multicolumn{4}{l}{\textit{Majority vote vs.\ deterministic checks}} \\
Executed & $100.0\%$ & $100.0\%$ & $100.0\%$ \\
Constraint satisfied & $100.0\%$ & $100.0\%$ & $100.0\%$ \\
Ground-truth match & $83.3\%$ & $100.0\%$ & $66.7\%$ \\
\midrule
\multicolumn{4}{l}{\textit{Ranking stability}} \\
Mean Spearman $\rho$ (10 repeat pairs) &
$0.924$ & \multicolumn{2}{c}{range $[0.862,0.976]$} \\
Optimal $>$ suboptimal (225 pairs) &
$83.3\%$ & \multicolumn{2}{c}{range $80.0$--$86.7\%$} \\
\bottomrule
\end{tabular}
\caption{Evaluator reliability on 30 fixed NLP4LP traces (15 optimal and
15 suboptimal), each graded five times with GPT-5 using the Appendix~A
evaluation prompt (150 calls total). Boolean agreement is the mean fraction
of repeated evaluations matching the per-trace majority label. Score
statistics are computed within each trace. Ranking statistics compare the
five score vectors over the same 30 traces.}
\label{tab:evaluator_reliability}
\end{table}

Because the scalar evaluator reward directly affects MCTS selection and backpropagation, we examine its repeatability under repeated grading. We select 30 fixed NLP4LP execution traces, consisting of 15 optimal and 15 suboptimal solutions; no failed-execution traces occurred in the sampled pool. Each trace is evaluated five times using the same GPT-5 evaluator and identical evaluation prompt, yielding 150 evaluator calls. The generated program and execution trace are held fixed across repetitions, so any variation reflects evaluator stochasticity rather than code-generation variation. 

As shown in Table~\ref{tab:evaluator_reliability}, the structured Boolean judgments are highly stable: execution-success labels agree across repeated evaluations in $100\%$ of cases, constraint-satisfaction labels in $99.3\%$, and ground-truth-match labels in $100\%$. Scalar rewards show greater variation. All optimal traces receive a score of $10$ in every repetition. For suboptimal traces, the mean within-trace standard deviation is $0.54$ and the mean score range is $1.38$; across all traces, these values are $0.27$ and $0.69$, respectively. The largest observed range for an individual trace is $6.2$. 

Despite this scalar variation, the relative ordering of traces is largely stable. Pairwise Spearman rank correlations across the five repeated evaluations have a mean of $\rho=0.924$, ranging from $0.862$ to $0.976$. Across the 225 optimal--suboptimal trace pairs, the optimal trace receives a strictly higher score in $80.0$--$86.7\%$ of comparisons across repeats, with a mean of $83.3\%$. Only 2 of the 30 traces cross the perfect-score boundary of $10$ across repetitions.

As a post-hoc check, we also compare the evaluator's majority-vote Boolean judgments with deterministic checks based on return code, timeout status, parsed numerical output, and tolerance-based comparison with the reference solution. Agreement is $100\%$ for execution success and constraint satisfaction and $83.3\%$ for ground-truth matching. All five ground-truth disagreements occur among the 15 suboptimal traces; in four cases, the evaluator assigns a score of $10$ in all five repetitions even though the deterministic numerical match fails.

Overall, the evaluator is highly repeatable for execution and constraint judgments and preserves the broad ordering of solution quality across repeated grading, while scalar rewards remain variable for some suboptimal traces. These results indicate that evaluator stochasticity is limited for the structured judgments but remains a source of uncertainty in the scalar reward signal.

\subsection{Equal Best-of-$N$ Comparison} 
\label{sec:equal-best-of-n}

\begin{table}[t]
\centering
\small
\begin{tabular}{lccccc}
\toprule
Method &
Mean Reward &
Best-of-3 Reward &
Mean Optimality &
Success@$3$ &
Avg.\ Tokens \\
\midrule
One-shot
& 9.09 & 9.52 & \textbf{75.0\%} & 78.0\% & \textbf{13{,}788} \\

MCTS-RDP
& 8.53 & \textbf{9.63} & 66.3\% & \textbf{82.0\%} & 64{,}584 \\

MCTS-RDP-10
& \textbf{9.16} & 9.60 & 74.0\% & 78.0\% & 19{,}595 \\
\bottomrule
\end{tabular}
\caption{Equal best-of-$N$ comparison on a random subset of 100 NLP4LP
problems with $N=3$. Success@$3$ denotes the fraction of problems for
which at least one of the three runs matches the reference solution.
The comparison equalizes the number of candidate opportunities rather
than total inference cost.}
\label{tab:equal_best_of_n}
\end{table}

The main experiments compare methods under their standard inference procedures, which differ in the number and structure of LLM generations. To examine whether the improvement of MCTS-RDP over one-shot generation is explained simply by considering multiple candidate solutions, we conduct an additional equal best-of-$N$ comparison on a fixed random subset of 100 NLP4LP problems, with $N=3$. 

For each problem, the one-shot baseline generates three independent solutions. Full MCTS-RDP performs three search simulations while maintaining a shared search tree for the problem. We also evaluate the search--then--apply variant under the same three-run evaluation protocol. For each method, we report the mean performance across the three runs and the best result obtained among them. 

Table~\ref{tab:equal_best_of_n} reports mean evaluator reward, best-of-three reward, mean optimality, Success@$3$ (the fraction of problems for which at least one of the three runs matches the reference solution), and average token usage. This experiment equalizes the number of candidate opportunities rather than total inference cost; token usage is therefore reported explicitly.

Under the equal best-of-three protocol, full MCTS-RDP attains the highest observed Success@$3$ ($82.0\%$ versus $78.0\%$ for one-shot generation) and the highest best-of-three reward ($9.63$ versus $9.52$), despite lower mean reward and mean optimality across its individual simulations. This indicates greater variability across MCTS simulations, but also a higher observed probability of finding a strong candidate when multiple search outcomes are considered. 

The Search-then-Apply variant substantially reduces this cost: average token usage drops from 64{,}584 to 19{,}595 while best-of-three reward remains similar ($9.60$ versus $9.63$). Its Success@$3$ is $78.0\%$, matching the one-shot baseline. Thus, full search provides the strongest observed best-of-$N$ performance, while search--then--apply offers a lower-cost trade-off.

\subsection{Analysis of Computational Cost}
\label{sec:cost}

\input{table4}

A key cost driver of MCTS-RDP is the number of LLM calls incurred during tree expansion, simulation, and optional refinement. While this added inference-time compute yields large gains in executability and solution quality, it is important to understand how much search is necessary in practice and whether search can be reduced across many instances.

\textbf{End-to-end inference time.} Tables~\ref{tab:nl4opt} and~\ref{tab:nlp4lp} report average end-to-end inference time. On modified NL4OPT with GPT-5, MCTS-RDP requires 100.26 seconds per instance, compared with 76.93 seconds for one-shot generation and 267.34 seconds for Chain-of-Thought. With Gemini-2.5-Pro, the corresponding times are 39.72, 37.17, and 134.01 seconds, respectively. 

A similar pattern is observed on NLP4LP. With GPT-5, MCTS-RDP requires 70.08 seconds, compared with 59.41 seconds for one-shot generation and 247.06 seconds for Chain-of-Thought. With Gemini-2.5-Pro, the corresponding times are 38.28, 33.19, and 124.26 seconds. Thus, MCTS-RDP introduces moderate overhead relative to one-shot generation but remains substantially faster than the evaluated Chain-of-Thought configuration.

\textbf{Amortizing search via Search-then-Apply.}
Table~\ref{table4} studies the method proposed in \ref{sta} under different amounts of active search. For the 100-instance modified NL4OPT benchmark, we retain the same ten sequential blocks of 10 problems and vary the number of SEARCH instances within each block. MCTS-RDP-$k$ denotes a configuration with $k$ SEARCH instances in total across the benchmark: $k=10,20,30,40,$ and $50$ correspond respectively to 1, 2, 3, 4, and 5 SEARCH instances per block, with the remaining instances processed in APPLY mode using the best block-local exemplar discovered so far. This search-fraction sweep is conducted as a separate experiment from the main NL4OPT evaluation in Table~\ref{tab:nl4opt}.

The results show that a relatively small amount of active search is sufficient to obtain strong performance. With MCTS-RDP-10, only one instance per block is assigned to SEARCH, yet the method achieves $100\%$ execution success, an average reward of $9.67$ (SD $1.19$), $98\%$ constraint satisfaction, and $96\%$ optimality, using 7,133 tokens on average. Increasing the number of SEARCH instances raises inference cost but produces comparatively modest changes in performance. Average token usage increases from 7,133 for MCTS-RDP-10 to 13,956 for MCTS-RDP-50, while execution success remains $100\%$ and optimality remains high across the configurations. These results indicate diminishing returns from increasing the fraction of actively searched instances within each block and support the use of a small SEARCH fraction to amortize tree search.

\input{table5}

\subsection{Ablation Studies}

To isolate the effect of Monte Carlo Tree Search in our pipeline, we evaluate two simplified variants on NLP4LP: (i) a one-shot baseline that synthesizes a full solver program in a single pass, and (ii) a decomposition-only variant that constructs a RDP (context, objective, constraints) and generates code accordingly, but does not explore alternatives via tree search.

Table~\ref{table5} summarizes the results. The decomposition-only variant maintains near-perfect executability, but solution quality drops noticeably: reward decreases and variance increases, and optimality attainment falls to 60.74\%. This indicates that decomposition structure alone is not sufficient to consistently produce high-quality formulations—selecting which plan steps to use matters.

Compared to the one-shot baseline, decomposition-only also incurs substantially higher cost (tokens and inference time) while achieving lower optimality. Overall, these results suggest that the primary gains of our full method come from searching over RDPs with execution-grounded feedback, rather than from decomposition alone.

\subsection{Failure Analysis on Hard Wireless Problems} The hard wireless benchmark contains 50 instances. MCTS-RDP generates executable programs for 35 instances and attains the reference optimum for 33 instances. Therefore, 15 instances fail during execution, while two additional instances execute successfully but do not attain the reference optimum. 

The hard cases are particularly challenging because the SINR constraints must be transformed into a solver-compatible form. Directly encoding the SINR ratio can violate disciplined convex programming requirements when optimization variables appear in both the numerator and denominator. 

Inspection of representative failed outputs revealed several recurring failure modes. The most common issue was the direct encoding of fractional SINR constraints in CVXPY, which violates disciplined convex programming rules when optimization variables appear in both the numerator and denominator. Other observed failures included incorrect algebraic reformulations of the SINR constraints, undefined variables or invalid mathematical operations in the generated script, and optimization parameters that were declared but not assigned numerical values before solving. These errors resulted in DCP violations, parameter errors, or other runtime failures. 

The two executable but non-optimal cases produced valid numerical outputs but did not recover the reference objective value. These results show that plan-level search substantially improves performance on difficult formulations, but it does not guarantee that every selected plan will be translated into correct and optimal solver code.

\subsection{Advantages and Limitations}

Compared with one-shot generation and Chain-of-Thought prompting, MCTS-RDP allocates inference-time computation to the formulation decisions that determine the variables, objective, constraints, and solver-compatible transformations. Unlike refinement methods that primarily repair an initial program, MCTS-RDP compares alternative formulation plans before selecting the final program. Its execution-grounded reward also allows search to account for executability, constraint satisfaction, and objective quality rather than relying only on the apparent quality of a natural-language reasoning trace. Finally, the search--then--apply variant can amortize the initial search cost when related instances share a transferable formulation structure. 

The approach also has several limitations. Full per-instance MCTS requires more LLM calls and tokens than one-shot generation and can introduce additional end-to-end inference time. Its search signal depends on a secondary LLM evaluator. Although our repeated-grading analysis shows that execution and constraint judgments are highly stable and that the broad ranking of solution quality is largely preserved across repeated evaluations, scalar rewards can vary for some suboptimal traces. Because these numerical rewards enter MCTS value estimates and backpropagation, evaluator stochasticity may influence individual search trajectories. In the current implementation, constraint satisfaction is also assessed by the evaluator rather than verified by a separate formal feasibility checker, while evaluation of optimality requires a reference or best-known solution. 

The score-based sibling-reuse rule is another approximation: it is a coarse branching-control heuristic rather than a semantic equivalence test. Equal evaluator scores may cause distinct plan steps to share a branch, while semantically similar steps may receive different scores and remain separate. More explicit semantic-similarity mechanisms could therefore improve branch management in future work. In addition, invalid plan steps are generally detected only after code realization and execution, so some search budget may be spent exploring inconsistent branches. The present experiments primarily cover linear and convex optimization formulations, and further evaluation is needed for mixed-integer, nonconvex, and broader domain-specific problems. Finally, the search--then--apply variant assumes that plans discovered during the search phase transfer effectively to subsequent instances from the same problem distribution.

%% file: table2.tex
\begin{table*}[!h]
\centering
\resizebox{\textwidth}{!}{%
\begin{tabular}{@{}llccccccc@{}}
\toprule\toprule
Method & Base Model 
& \shortstack{Code Execution\\Success Rate}
& \shortstack{Reward (Out of 10)\\$\pm$ SD}
& \shortstack{Constraint \\ Satisfied} 
& \shortstack{Optimality \\ Attained}
& \shortstack{Avg Tokens \\ Used}
& \shortstack{Avg End-to-End \\ Time (s)}\\
\midrule

\multirow{2}{*}{$\textbf{MCTS\text{-}RDP}$} 
    & GPT-5 
        & \textbf{100.00\%} & \textbf{9.56 $\pm$ 1.41} & \textbf{100.00\%} & \textbf{92.00\%} & 7517 & 100.26\\
    & Gemini-2.5-Pro 
        & \textbf{100.00\%} & \textbf{9.59 $\pm$ 1.46} & \textbf{98.00\%} & \textbf{91.00\%} & 7247 & 39.72 \\
\midrule

\multirow{2}{*}{$\text{Optimus}$} 
    & GPT-5 
        & 80.00\% & 7.25 $\pm$ 4.44 & 76.00\% & 69.00\% & 32130 & 350.72 \\
    & Gemini-2.5-Pro 
        & 90.00\% & 7.69 $\pm$ 4.02 & 82.00\% & 66.00\% & 43298 & 272.26  \\
\midrule

\multirow{2}{*}{$\text{Self-Refine}$} 
    & GPT-5 
        & 100.00\% & 9.31 $\pm$ 1.52 & 97.00\% & 77.00\% & 5722 & 80.42 \\
    & Gemini-2.5-Pro 
        & 100.00\% & 9.52 $\pm$ 1.48 & 97.00\% & 79.00\% & 6550 & 42.05 \\
\midrule

\multirow{2}{*}{$\text{Chain-of-Thought}$} 
    & GPT-5 
        & 83.00\% & 6.61 $\pm$ 4.30 & 72.00\% & 50.00\% & 18664 & 267.34 \\
    & Gemini-2.5-Pro 
        & 100.00\% & 7.09 $\pm$ 3.83 & 77.00\% & 58.00\% & 20517 & 134.01 \\
\midrule

\multirow{2}{*}{$\text{Baseline}$} 
    & GPT-5 
        & 98.00\% & 8.98 $\pm$ 2.17 & 93.00\% & 73.00\% & 5439 & 76.93 \\
    & Gemini-2.5-Pro 
        & 96.00\% & 9.12 $\pm$ 2.53 & 94.00\% & 74.00\% & 6319 & 37.17 \\
\bottomrule\bottomrule
\end{tabular}%
}
\caption{
Evaluation of \textbf{MCTS-RDP-10} (search--then--apply variant with a $10\%$ SEARCH fraction per
block) and baselines on the modified NL4OPT dataset using GPT-5 and Gemini-2.5-Pro. We report success rate, reward (mean $\pm$ standard deviation), constraint satisfaction rate, optimality attainment, average tokens used, and average end-to-end inference time (in seconds).
}
\label{tab:nl4opt}
\end{table*}

%% file: table3.tex
\begin{table*}[!t]
\centering
\resizebox{\textwidth}{!}{%
\begin{tabular}{@{}llccccccc@{}}
\toprule\toprule
Method & Base Model 
& \shortstack{Code Execution\\Success Rate}
& \shortstack{Reward (Out of 10)\\$\pm$ SD}
& \shortstack{Constraint \\ Satisfied} 
& \shortstack{Optimality \\ Attained}
& \shortstack{Avg Tokens \\ Used}
& \shortstack{Avg End-to-End \\ Time (s)}\\
\midrule

\multirow{2}{*}{$\textbf{MCTS\text{-}RDP}$} 
    & GPT-5 
        & \textbf{100.00\%} & \textbf{9.47 $\pm$ 1.37} & \textbf{97.52\%} & \textbf{78.10\%} & 7158 & 70.08 \\
    & Gemini-2.5-Pro 
        & \textbf{100.00\%} & \textbf{9.51 $\pm$ 1.59} & \textbf{97.52\%} & \textbf{76.45\%} & 6940 & 38.28 \\
\midrule

\multirow{2}{*}{$\text{Optimus}$} 
    & GPT-5
        & 72.73\% & 6.09 $\pm$ 4.81 & 66.53\% & 55.79\% & 29273 & 328.00 \\
    & Gemini-2.5-Pro 
        & 92.56\% & 8.41 $\pm$ 3.53 & 88.02\% & 69.42\% & 38182 & 244.68  \\
\midrule

\multirow{2}{*}{$\text{Self-Refine}$} 
    & GPT-5 
        & 100.00\% & 9.40 $\pm$ 1.52 & 97.93\% & 78.10\% & 4527 & 66.57 \\
    & Gemini-2.5-Pro 
        & 100.00\% & 9.44 $\pm$ 1.80 & 94.21\% & 74.38\% & 6333 & 37.57 \\
\midrule

\multirow{2}{*}{$\text{Chain-of-Thought}$} 
    & GPT-5 
        & 83.88\% & 6.60 $\pm$ 4.32 & 78.10\% & 54.13\% & 17012 & 247.06 \\
    & Gemini-2.5-Pro 
        & 99.59\% & 7.32 $\pm$ 3.77 & 78.93\% & 53.72\% & 19759 & 124.26 \\
\midrule

\multirow{2}{*}{$\text{Baseline}$} 
    & GPT-5 
        & 98.76\% & 9.21 $\pm$ 1.97 & 97.52\% & 76.03\% & 4545 & 59.41 \\
    & Gemini-2.5-Pro 
        & 98.35\% & 9.36 $\pm$ 2.25 & 95.04\% & 75.62\% & 5741 & 33.19\\
\bottomrule\bottomrule
\end{tabular}%
}
\caption{
Evaluation of \textbf{MCTS-RDP-10} (search--then--apply variant with a nominal $10\%$ SEARCH fraction per
block) and baselines on the NLP4LP dataset using GPT-5 and Gemini-2.5-Pro. We report success rate, reward (mean $\pm$ standard deviation), constraint satisfaction rate, optimality attainment, average tokens used, and average end-to-end inference time (in seconds).
}
\label{tab:nlp4lp}
\end{table*}

%% file: table1.tex
\begin{table*}[!t]
\centering
\resizebox{\textwidth}{!}{%
\begin{tabular}{@{}c*{4}{cc}@{}}
\toprule\toprule
{Method} 
& \multicolumn{2}{c}{Code Execution Success Rate} 
& \multicolumn{2}{c}{Reward (Out of 10) $\pm$ SD} 
& \multicolumn{2}{c}{Optimality Attained} 
& \multicolumn{2}{c}{Avg Tokens Used} \\
\cmidrule(lr){2-3} \cmidrule(lr){4-5} \cmidrule(lr){6-7} \cmidrule(lr){8-9}
Problem Difficulty 
& Easy & Hard 
& Easy & Hard 
& Easy & Hard 
& Easy & Hard \\
\midrule

$\textbf{MCTS-RDP}$
& \textbf{98.00\%} & \textbf{70.00\%}
& \textbf{9.11 $\pm$ 0.43} & \textbf{6.75 $\pm$ 3.89}
& \textbf{92.00\%} & \textbf{66.00\%}
& 67,825 & 171,915 \\

\midrule

$\text{Self-Refine}$
& 90.00\% & 18.00\%
& 8.66 $\pm$ 3.26 & 0.10 $\pm$ 1.89
& 78.00\% & 0.00\%
& 1,242 & 4,416 \\

\midrule

$\text{Chain-of-Thought}$
& 92.00\% & 14.00\%
& 5.78 $\pm$ 4.38 & -0.30 $\pm$ 1.67
& 54.00\% & 4.00\%
& 3,077 & 4,755 \\

\midrule

$\text{GPT-4 Baseline}$
& 72.00\% & 0.00\%
& 4.75 $\pm$ 1.29 & -1.00 $\pm$ 0.00
& 58.00\% & 0.00\%
& 736 & 525 \\

\midrule

$\text{GPT-3.5-turbo Baseline}$
& 76.00\% & 0.00\%
& 4.36 $\pm$ 4.35 & -1.00 $\pm$ 0.00
& 42.00\% & 0.00\%
& 768 & 552 \\

\bottomrule\bottomrule
\end{tabular}%
}
\caption{
Performance of MCTS-RDP (full version) versus baselines on 100 network optimization tasks (50 easy, 50 hard). We report execution success rate, reward (mean $\pm$ standard deviation), optimality attainment, and average token usage.
}
\label{table1}
\end{table*}

%% file: table4.tex
\begin{table*}[ht]
\centering
\resizebox{\textwidth}{!}{%
\begin{tabular}{@{}lcccccc@{}}
\toprule\toprule
Method
& \shortstack{Code Execution\\Success Rate}
& \shortstack{Reward (Out of 10)\\$\pm$ SD}
& \shortstack{Constraint\\Satisfied}
& \shortstack{Optimality\\Attained}
& \shortstack{Avg Tokens\\Used}\\
\midrule

$\text{MCTS-RDP-10}$
    & 100.00\% & 9.67 $\pm$ 1.19 & 98.00\%  & 96.00\% & 7133 \\ 
\midrule
$\text{MCTS-RDP-20}$
    & 100.00\% & 9.58 $\pm$ 1.29 & 98.00\%  & 96.00\% & 8956 \\ 
\midrule
$\text{MCTS-RDP-30}$
    & 100.00\% & 9.61 $\pm$ 1.24 & 100.00\% & 96.00\% & 10215 \\ 
\midrule
$\text{MCTS-RDP-40}$
    & 100.00\% & 9.32 $\pm$ 1.53 & 98.00\%  & 87.00\% & 12195 \\ 
\midrule
$\text{MCTS-RDP-50}$
    & 100.00\% & 9.65 $\pm$ 1.00 & 98.00\%  & 95.00\% & 13956 \\ 
\bottomrule\bottomrule
\end{tabular}%
}

\caption{
\textbf{Generalization of MCTS-RDP from Partial Search.}
We evaluate the Search-then-Apply procedure on modified NL4OPT. The 100 instances are divided into ten sequential blocks of 10 problems. MCTS-RDP-$k$ denotes $k$ SEARCH instances in total, corresponding to $k/10$ SEARCH instances per block for $k\in\{10,20,30,40,50\}$; the remaining instances in each block are processed in APPLY mode using the best block-local exemplar. Increasing the SEARCH fraction raises token cost while yielding relatively modest changes in performance.}

\label{table4}
\end{table*}

%% file: table5.tex
\begin{table*}[!ht]
\centering
\resizebox{\textwidth}{!}{%
\begin{tabular}{@{}lccccccc@{}}
\toprule\toprule
Method
& \shortstack{Code Execution\\Success Rate}
& \shortstack{Reward Avg\\(Out of 10)}
& \shortstack{Constraint \\ Satisfied}
& \shortstack{Optimality \\ Attained}
& \shortstack{Avg Tokens \\ Used}
& \shortstack{Avg End-to-End \\ Time (s)}\\
\midrule
\textbf{MCTS-RDP (search-then-apply)} 
& 100.00\% & 9.47 $\pm$ 1.37 & 97.52\% & 78.10\% & 7158  & 70.08 \\
\addlinespace[0.08cm]
~~w/o MCTS (Decomposition-only)
& 99.59\% & 8.28 $\pm$ 2.73 & 95.45\% & 60.74\% & 20359 & 297.98 \\
~~w/o MCTS + w/o decomposition (one-shot)
& 98.76\% & 9.21 $\pm$ 1.97 & 97.52\% & 76.03\% & 4545  & 59.41 \\
\bottomrule\bottomrule
\end{tabular}%
}
\caption{
Ablation on NLP4LP (GPT-5). We compare the reported MCTS-RDP search-then-apply configuration
against a decomposition-only variant without tree search and a one-shot baseline. Metrics include execution success, reward, constraint satisfaction, optimality attainment (ground-truth match), token usage, and average end-to-end inference time (in seconds).
}
\label{table5}
\end{table*}

%% file: 06conclusion.tex
\section{Conclusion}
In this work, we introduce MCTS-OPS, a novel framework that integrates large language models (LLMs) and Monte Carlo Tree Search (MCTS) by treating prompt selection as a planning problem. Rather than relying on static or human-generated prompts, MCTS-OPS dynamically constructs prompt sequences through tree-based exploration. It enables adaptive decomposition and synthesis of problem-solving strategies. At each step, MCTS searches and explores candidate prompts and generates code based on the prompts. The code is then executed and evaluated for feasibility and optimality.  

We conduct extensive experiments on a set of 100 optimization tasks covering domains such as wireless communication, convex optimization, and signal processing. Our results show that MCTS-OPS significantly outperforms strong LLM baselines in terms of success rate, reward quality, stability, and optimality. Notably, MCTS-OPS achieves a 25-70 percentage point increase in success rate, 2$\sim$4$\times$  higher reward and 3$\times$ lower standard deviation, and up to 10$\times$ improvement on optimality attainment.

This study demonstrates the potential of integrating classical planning algorithms with generative models for complex problem-solving and planning, and highlights new opportunities for research on strategic decision-making with LLMs. Our work lays the foundation of future research on planning-aware LLMs, interpretable prompt generation, and generalizable solvers for structured decision problems.

\section{Limitations}
While our current evaluation focuses on network optimization problems with corresponding objectives and constraints, the proposed framework can potentially be extended to a broader class of optimization tasks and planning problems. Future work may explore its applicability to combinatorial optimization, decision-making under uncertainty, and real-world planning domains to further assess its generalization capabilities across diverse problem types.

%% file: appendix.tex
\appendix
\section{LLM Prompt Templates}
\label{app:system-prompts}

This appendix documents the role-specific system templates used in our \textbf{MCTS-RDP} pipeline.
We use a modular set of templates aligned with
distinct functional roles: (i) decomposing the natural-language problem into atomic elements,
(ii) proposing candidate \emph{reasoning decomposition plan steps}, (iii) scoring candidate steps for
clarity and downstream code correctness, (iv) synthesizing solver-ready code incrementally using a
code prefix, (v) assigning execution-grounded rewards from runtime traces, and (vi) optionally repairing
faulty programs using structured feedback. 


\begin{figure}[H]
\centering
\small
\begin{tcolorbox}[colback=blue!5, colframe=blue!60!black, title=User Input (Natural-Language Optimization Problem), fonttitle=\bfseries]
\textbf{Problem:} \{PROBLEM\_DESCRIPTION\}
\end{tcolorbox}
\end{figure}

\begin{figure}[H]
\centering
\small
\begin{tcolorbox}[colback=orange!5, colframe=orange!90!black, title= Prompt for Decomposition Model, fonttitle=\bfseries]
You are a systems engineer who breaks optimization problems into structured, descriptive sentences.
Break down the following optimization problem into 3 concise descriptive sentences that describe first the context, then the objective, and last the constraints. Each sentence must end with a newline character:

\{PROBLEM\_DESCRIPTION\}
\end{tcolorbox}
\end{figure}

\begin{figure}[H]
\centering
\small
\begin{tcolorbox}[colback=orange!5, colframe=green!60!black, title=Prompt for Generating a Plan-Step Instruction, fonttitle=\bfseries]
You are an expert assistant for Python optimization code generation.
Transform a single sentence from an optimization problem into a concise, precise instruction
that another LLM can use to generate Python code with an appropriate optimization library.

\textbf{Library consistency:}
\begin{itemize}
\item If depth = 0: Choose any suitable optimization library (e.g., PuLP, Pyomo, OR-Tools, CVXPY, mip) and stick with it for subsequent steps.
\item If depth $>$ 0: Reuse the same optimization library already used in prior steps and keep variable names/structure; do not switch libraries.
\end{itemize}

Return only the instruction string (no explanations, code, or reasoning).

The following is a sentence from an optimization problem:

"\{SENTENCE\}"

\textbf{If Intermediate Step:}

Write a single, high-quality instruction that directs an LLM to generate or modify Python code to reflect \emph{only} this sentence
(e.g., define variables, parameters, a constraint, or an objective) using the same optimization library as prior steps.
Do \textbf{not} solve the model or print results at this stage; produce only the incremental change for this sentence.
Return only the instruction string.

\textbf{If Final Step:}

Write a single, high-quality instruction that directs an LLM to produce a complete, runnable Python program that assembles the entire optimization model using the same optimization library as prior steps.
Import required libraries, define all parameters, variables, objective, and constraints, execute the optimization, and print the final decision variable values and the objective value.
If a single numeric result is requested, also print:
\quad FINAL\_ANSWER: <number>
\end{tcolorbox}
\end{figure}

\begin{figure}[H]
\centering
\small
\begin{tcolorbox}[colback=orange!5, colframe=red!60!black, title=Prompt for Plan-Step Scorer, fonttitle=\bfseries]
You score instructions based on descriptiveness and cohesion.

Rate this instruction from 1--10 for clarity and detail:

\{STEP\_INSTRUCTION\}

Only return a number.
\end{tcolorbox}
\end{figure}

\begin{figure}[H]
\centering
\small

\begin{tcolorbox}[colback=orange!5, colframe=purple!60!black,
  title=Prompt for Code Generation Model (No Prior Code),
  fonttitle=\bfseries]
You are a Python optimization expert. Generate complete, runnable Python code using any suitable optimization library.

Requirements:
\begin{itemize}
\item Import all required libraries and define all variables, parameters, objectives, and constraints.
\item Ensure the formulation is valid and can be solved by the chosen library.
\item Assign values to all parameters before solving.
\item Solve the optimization problem and print key variable values and the final objective value.
\item If a single numeric final quantity is requested, also print:
\quad FINAL\_ANSWER: <number>
\item Do not include explanations or comments.
\end{itemize}

Instruction:
\{STEP\_INSTRUCTION\}
\end{tcolorbox}

\begin{tcolorbox}[colback=orange!5, colframe=purple!60!black,
  title=Prompt for Code Generation Model (Update Existing Code),
  fonttitle=\bfseries]
You are a Python optimization expert. Modify the existing optimization code according to the new instruction.

Requirements:
\begin{itemize}
\item Preserve the overall structure and variable naming of the existing code.
\item Make only the necessary edits to incorporate the new instruction.
\item Reuse the optimization library already used in the code (do not switch libraries).
\item Ensure all parameters and variables are correctly defined and assigned before solving.
\item Ensure the resulting code is complete, runnable, and includes all necessary imports.
\item Solve the optimization problem and print key variable values and the final objective value.
\item If a single numeric final quantity is requested, also print:
\quad FINAL\_ANSWER: <number>
\item Do not include explanations or comments.
\end{itemize}

Existing Code:
\{EXISTING\_CODE\}

Instruction:
\{STEP\_INSTRUCTION\}

Output only the modified full Python code.
\end{tcolorbox}

\label{fig:code-synth-prompts}
\end{figure}

\begin{figure}[H]
\centering
\small
\begin{tcolorbox}[colback=orange!5, colframe=black!75!black, title=Prompt for Execution-Grounded Evaluator, fonttitle=\bfseries]
You are a strict grader for optimization code. Return ONLY a single JSON object with keys:
score (float in [-1,10]), explanation (string), executed (bool),
\text{constraint\_satisfied} \;(\text{bool}),\quad \text{matches\_ground\_truth} \;(\text{bool})

Do not add any text outside JSON.

Problem description:
\{PROBLEM\_DESCRIPTION\}

Execution trace (stdout + stderr):
\{EXEC\_TRACE\}

(Optional) Ground-truth numeric answer (if applicable): \{GROUND\_TRUTH\}
Numeric tolerance for matching (absolute): \{TOLERANCE\}
If a final numeric answer can be extracted from the execution trace, compute |pred - ground\_truth|
and set matches\_ground\_truth = true if the difference is $\le \{TOLERANCE\}$, else false.

Scoring rubric:
\begin{itemize}
\item If the program did not execute (timeout, runtime error, syntax error), set score = -1 and executed=false.
\item Otherwise, start from a base score and increase for:
(i) executability and completeness of results,
(ii) constraint satisfaction (feasibility),
(iii) objective quality and plausibility,
(iv) (optional) closeness to ground-truth if provided.
\item Keep the final score in [-1, 10].
\item If no meaningful result can be extracted despite successful execution, you may assign a low nonnegative score.
\end{itemize}

Return strictly JSON, e.g.:\\
\{
  "score": 7.5,\\
  "explanation": "why",\\
  "executed": true,\\
\texttt{constraint\_satisfied: false,}\\
\text{matches\_ground\_truth: false}\\
\}
\end{tcolorbox}

\centering
\small
\begin{tcolorbox}[colback=orange!5, colframe=teal!70!black, title=Prompt for Optional Refinement Loop, fonttitle=\bfseries]
You are a helpful Python assistant that improves optimization code.

The following code was generated to solve the optimization problem:

\{PROBLEM\_DESCRIPTION\}

Here is the original Python code:
\{CODE\}

However, the code did not perform well. Here's the feedback:
\{FEEDBACK\}

Please revise and improve the code accordingly, addressing the issues mentioned in the feedback.
Make sure all variables and constraints align with the problem description.
Return ONLY the updated Python code; do not provide any explanation.
\end{tcolorbox}
\end{figure}

\section{Experimental Configuration and Reproducibility}
\label{app:reproducibility}

\subsection{Experimental Configuration}

All GPT-5 experiments use the OpenAI Chat Completions API with model
\texttt{gpt-5} and sampling temperature $1$, while Gemini experiments use
\texttt{gemini-2.5-pro} at temperature $1$. Generated programs are executed
in a separate \texttt{python3} subprocess with a $30$\,s timeout. Standard
output and standard error are concatenated into the execution trace supplied
to the evaluator. The original API-based experiments do not use a fixed
random seed. Runs were conducted on Linux with Python~$3.10+$, an AMD
EPYC~7513 CPU, and an NVIDIA RTX~A6000 GPU.

The evaluator uses the prompt in \ref{app:system-prompts} and an
absolute numerical matching tolerance of $\tau=10^{-4}$. Ground truth is
not included in the initial code-generation prompt, but is provided to the
post-execution evaluator. When self-refinement is triggered, the refinement
model receives the evaluator's textual feedback; consequently,
reference-informed information may indirectly affect refinement. The evaluator does not receive the generated program source code directly; its inputs are the problem description, execution trace, and, when available, the numerical reference value and matching tolerance.

The principal MCTS settings are $c=\sqrt{3}$ for UCT and three simulations
per SEARCH instance. Self-refinement permits at most two retries and is
triggered after execution failure or when the evaluator reward is below
$1$. The implementation also uses the score-based sibling-reuse heuristic
described in Section~\ref{sec:decomposition-planstep}.

For the main GPT-5 NLP4LP search--then--apply experiment, the implementation
uses blocks of $B=24$ and $k_b=3$ SEARCH instances per full block. The
archived $n=242$ run therefore contains 32 SEARCH and 210 APPLY instances,
where APPLY is labeled \texttt{EXAMPLES\_ONLY} in the archived JSON.
Additional details of the search--then--apply procedure are given in
Section~\ref{sta}.

A separate equal best-of-$N$ study
(Section~\ref{sec:equal-best-of-n}) uses a fixed random subset of 100
NLP4LP problems selected with seed~$42$. This auxiliary experiment is
separate from the protocol used for the main NLP4LP tables.

\begin{table}[t]
\centering
\caption{Principal implementation settings used in the released GPT-5
experiments.}
\label{tab:exp-config}
\small
\begin{tabular}{ll}
\toprule
Setting & Value \\
\midrule
GPT-5 temperature & $1$ \\
Gemini temperature & $1$ \\
Execution timeout & $30$\,s \\
Ground-truth tolerance $\tau$ & $10^{-4}$ \\
UCT exploration constant $c$ & $\sqrt{3}$ \\
MCTS simulations per SEARCH instance & $3$ \\
Maximum self-refinement retries & $2$ \\
\bottomrule
\end{tabular}
\end{table}

\subsection{Metric Definitions and Aggregation}
The released JSON files store problem-level quantities that are aggregated
by \texttt{aggregate\_results.py}. Dataset-level execution success,
constraint satisfaction, optimality attainment, token usage, and
end-to-end time are computed as unweighted means over the corresponding
problem-level fields. Reward is reported as the mean and sample standard
deviation of the problem-level evaluator scores.

Specifically, the aggregation uses
\texttt{success\_rate},
\texttt{avg\_score},
\texttt{constraint\_satisfaction\_rate},
\texttt{ground\_truth\_match\_rate},
\texttt{avg\_tokens}, and
\texttt{avg\_time}.
Optimality attainment corresponds to the stored \texttt{ground\_truth\_match\_rate}, i.e., the LLM evaluator's \texttt{matches\_ground\_truth} judgment. The evaluator prompt instructs the model to use the numerical tolerance above; this should not be interpreted as an independently computed deterministic residual check.

For MCTS SEARCH instances, three simulations share one search tree and the
reported problem-level result corresponds to the selected candidate from
those simulations. APPLY instances use the standard search--then--apply
generation procedure described in Section~\ref{sta}.

Compatible archived results can be summarized using
\texttt{python3 aggregate\_results.py --agg best}.

\subsection{Released Experimental Artifacts} \label{app:artifacts} 

The released repository contains both complete benchmark runs and partial archived snapshots. Complete artifacts are used for table reconstruction where available; partial files are retained for transparency but should not be interpreted as full-benchmark results. In particular, the released GPT-5 and Gemini NLP4LP MCTS logs contain all 242 instances, whereas some baseline artifacts cover only subsets of the corresponding benchmark. The repository README provides the file-to-experiment mapping and indicates whether each released artifact is complete, partial, or auxiliary.